\documentclass{article}

\usepackage{microtype}
\usepackage{graphicx}
\usepackage{subfigure}
\usepackage{booktabs} 
\usepackage{amsmath}
\usepackage{hyperref}
\usepackage{multirow}

\usepackage[accepted]{mlsys2025}

\mlsystitlerunning{Arbor: How Search Policy Shapes Agent Behavior in System Optimization}

\begin{document}

\twocolumn[
\mlsystitle{Arbor: How Search Policy Shapes Agent Behavior in System Optimization}



\mlsyssetsymbol{equal}{*}

\begin{mlsysauthorlist}
\mlsysauthor{Neha Prakriya}{}
\mlsysauthor{Chaojun Hou}{}
\mlsysauthor{Zheng Gong}{}
\mlsysauthor{Huasha Zhao}{}
\mlsysauthor{Yanyuan Qin}{}
\mlsysauthor{Chris Cai}{}
\mlsysauthor{Xi Zhao}{}
\mlsysauthor{Zhenyu Gu}{}
\mlsysauthor{Emad Barsoum}{}
\end{mlsysauthorlist}

\begin{center}
    AMD
\end{center}
\mlsyscorrespondingauthor{Neha Prakriya}{neha.prakriya@amd.com}


\vskip 0.3in

\begin{abstract}
Achieving peak performance on modern accelerators requires coordinated optimization across kernels, operator fusion, KV-cache management, quantization, scheduling, and communication overlap, a process that demands weeks of expert effort per model and hardware platform. We present Arbor, a system that gives autonomous agents an explicit optimization tree to reason over, backtrack through, and deepen with experience, with pluggable search policies, scored bandit traversal, parallel agent swarms, and depth-first sweep, sharing a common harness of containerized isolation, automatic rollback, and end-to-end validation. Arbor generalizes across domains: +40\% to +193\% on six production inference models across two GPU generations, +42.8\% on LLM training, and +41\% on recommendation training, where agents autonomously wrote kernel-level and framework-level optimizations spanning thousands of lines of code. In controlled ablations where every search policy receives the same agent, tree, and tools, we find that search structure shapes both the magnitude and the nature of the optimizations. The bandit achieves +30.1\% through targeted framework fixes, the sweep +26.4\% through deep kernel rewrites, and the swarm +33.4\% by compounding independent changes. An agent-driven loop that receives the same harness, tools, and information, but selects its own targets reaches +14\%, writing 4$\times$ more code per percentage point of improvement than Arbor's structured search policies. We additionally introduce an evaluation framework for autonomous loop-based agents spanning efficiency, grounding, robustness, and depth, designed to characterize agent search behavior across long-running optimization campaigns. 
\end{abstract}
]
\section{Introduction}
\label{sec:intro}

Autonomous agents operating in write-evaluate-improve loops have 
demonstrated strong results on targeted optimization tasks, from 
discovering improved algorithms~\cite{alphaevolve2025} to 
producing GPU kernels that exceed vendor library 
throughput~\cite{chen2026avo,duodiscover2026}. These systems optimize 
a single target against a single evaluator within a single campaign. 

Full-system optimization poses a fundamentally harder problem. The 
stack spans kernels, operator fusion, memory management, scheduling, 
and communication overlap, with many candidate targets at each layer 
and a multi-dimensional objective where throughput, latency, and cost 
trade off against each other. Changes at one layer affect others in 
ways no component test can predict. For example, in one of our 
campaigns, an attention kernel rewrite that runs 2$\times$ faster in 
isolation regresses end-to-end throughput by 38\% under production 
traffic because it changes the shape seen by a downstream kernel 
tuned for the wider layout (Section~\ref{sec:validation}). The optimization process demands weeks of expert effort per model and hardware 
platform, making it a bottleneck in new hardware bring-up.

This raises the first question:

\begin{quote}
\emph{How should an agent navigate a large, multi-layered optimization space where candidates range from kernel rewrites to framework configuration, changes at one layer affect others, and each evaluation is expensive?}
\end{quote}


\noindent A natural approach is to let a coding agent drive the 
process end-to-end: profile the system, choose a bottleneck, 
implement a fix, measure, and repeat. In practice, when the agent 
controls both \emph{what} to optimize and \emph{how}, it gravitates 
toward quick wins, oscillates between targets when the current one 
gets difficult, and abandons promising directions after a single 
failure instead of persisting through the multi-iteration effort 
required for deep kernel rewrites. The result is broad but shallow 
exploration that settles in local minima rather than compounding 
gains across the stack. Sustaining optimization over days or weeks 
requires a well-designed harness around the agent that separates 
critical responsibilities: containerized isolation so successive 
changes do not accumulate in a dirty workspace, automatic rollback 
so a crashing change does not stall the campaign, persistent state 
so the agent does not lose track of prior outcomes across context 
windows, and harness-controlled validation so the agent cannot 
inadvertently modify the benchmark it is measured against.

\textbf{Navigating the search space.} We present Arbor, a system 
that separates three components: the \emph{search policy}, the 
\emph{execution infrastructure}, and the \emph{task definition} 
(Figure~\ref{fig:arch}). An \emph{optimization tree} of scored 
candidates gives the agent an explicit map of the search space to 
reason over, backtrack through, and deepen with experience. The key 
design decision is who picks the next target: in Arbor's structured 
search policies, the \emph{harness} selects the target based on the 
tree's accumulated evidence, pushing the agent to commit to 
difficult optimizations rather than retreating to easy ones. 
Pluggable policies determine the selection strategy: a scored bandit 
that adapts as outcomes accumulate, a swarm that dispatches parallel 
agents to compound independent changes, or a depth-first sweep that 
sustains focus within a single category. The execution 
infrastructure isolates each attempt in a containerized workspace 
with git-based rollback, end-to-end validation, and persistent 
state across context windows. The task definition encapsulates a 
domain: the target system, the workload, the benchmark that measures 
progress, and the correctness checks that gate every change. 
Swapping the task definition moves Arbor from LLM inference to 
training to recommendation-model optimization with no change to the 
agent or harness.

Given multiple search policies operating over the same space, a second question arises:

\begin{quote}
\emph{How do we evaluate the search policy itself, not just the final outcome?}
\end{quote}

\noindent A harness-driven policy and an agent-driven loop may both reach +30\%, yet differ in token cost, code efficiency, coverage of the landscape, and whether independent runs converge.

\textbf{Evaluating the search policy.} Recent work recognizes that evaluating agents requires more than pass/fail: Harness-Bench~\cite{harnessbench2026} shows that the same model's ranking can reverse by over 30 points depending on the harness, and TRACE~\cite{trace2026} evaluates trajectory efficiency for research agents. These frameworks target single-task completion or short trajectories. System optimization campaigns run for hundreds of iterations over days, and the search space evolves as bottlenecks shift. We define metrics across four dimensions, efficiency, grounding, robustness, and depth, that capture how search policy shapes not only the performance gained but the code the agent produces: whether it commits to hard optimizations or skims easy ones, writes targeted kernel rewrites or speculative configuration changes, and whether independent runs converge.

The contributions are:

\begin{enumerate}
\item \textbf{Arbor}, a search infrastructure that separates the task definition (what to optimize) from the search policy (how to traverse) from the execution infrastructure (isolation, validation, rollback), enabling controlled comparison of search policies under identical agent, tools, and hardware.

\item \textbf{An evaluation framework} spanning four dimensions and ten metrics: efficiency (improvement per token, per GPU-hour, per iteration, and how decisively the agent pursues each target), grounding (measurement integrity, investigation before editing), robustness (recovery from failures, diagnosis before abandonment), and depth (fraction of attempts that survive validation, coverage of the search space).

\item \textbf{Results across three domains}: LLM inference (+40--193\% on six production models, two GPU generations), LLM training (+42.8\%), and recommendation training (+41\%), with a controlled ablation showing that the bandit produces 4$\times$ fewer lines of code per percentage point of improvement than agent-driven search with low variance (CV$=$0.01 vs.\ 0.34), and that depth-first traversal pushes the agent into deep kernel rewrites that the agent-driven baseline never attempts.
\end{enumerate}

Section~\ref{sec:design} describes the Arbor architecture, Section~\ref{sec:eval} introduces the evaluation framework, Section~\ref{sec:experiments} presents results and policy analysis across all three domains, and Section~\ref{sec:related} discusses related work.

\section{System Design}
\label{sec:design}

\begin{figure*}[t]
\centering
\includegraphics[width=\textwidth]{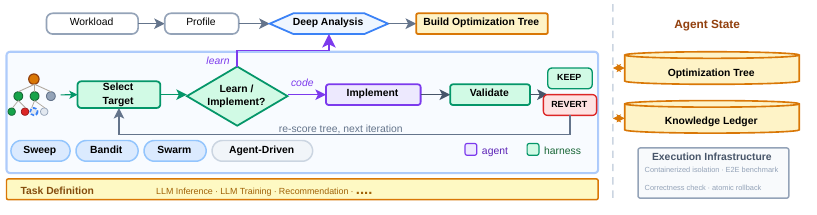}
\caption{Arbor architecture. Top: an initialization phase profiles 
the workload and constructs the optimization tree. Center: the 
iteration loop, where the harness selects a target, decides between 
a learn or implement iteration, and validates the result 
(green = harness, purple = agent). Pluggable policies (Sweep, 
Bandit, Swarm, Agent-Driven) control target selection. Right: 
persistent agent state (optimization tree and knowledge ledger) 
shared across all policies and iterations. Bottom: the task 
definition specifies the target system and benchmark; the execution 
infrastructure provides containerized isolation, end-to-end 
validation, and atomic rollback.}

\label{fig:arch}
\end{figure*}

Arbor separates three components: the \emph{task definition} specifies what to optimize and how to measure it, the \emph{search policy} determines how to traverse the optimization space, and the \emph{execution infrastructure} isolates, validates, and persists each attempt (Figure~\ref{fig:arch}). All four search policies presented in this paper, including the agent-driven baseline, share the same execution infrastructure and agent state; only the target selection rule differs.

This section describes each component, starting with the optimization problem itself, then the state over which all our policies reason, the validation that keeps the search honest, and the policies that traverse it.

\subsection{The Optimization Problem}
\label{sec:formulation}

We use DeepSeek-V4.1-Flash inference on 4$\times$~MI355X (TP$=$4) as the running example throughout this paper, benchmarked on AgentX~\cite{inferencex}, an open-source agentic inference benchmark derived from 393 real Claude Code sessions. AgentX replays the traffic \emph{shape} of real coding-agent interactions, including multi-turn contexts at up to 1M tokens, subagent fan-out (44\% of sessions), prefix sharing, and inter-turn timing, exercising the full serving stack from KV-cache management to scheduling under concurrent load. Arbor starts by profiling this workload and analyzing execution traces, constructing a tree of optimization categories; for DSv4.1-Flash this yields an initial set of 5~categories and 
17~leaf targets spanning kernels, operator fusion, memory management, 
quantization, and scheduling; we present the tree and how different policies 
traverse it in Section~\ref{sec:exp-inference} 
(Figure~\ref{fig:coverage}). Each iteration follows a think$\to$implement$\to$validate cycle: the harness provides the agent with a target, the agent analyzes the current state, writes and deploys a code change, then the harness restarts the target system, runs the full benchmark, and verifies correctness. The benchmark alone takes up to an hour, making exhaustive evaluation infeasible.

The optimization landscape also shifts as the search progresses. In one DSv4.1-Flash session, the agent applies a sequence of changes across different categories: a Triton kernel fallback fix, an FP8 scale-layout correction, a fused operator kernel, and finally tuned GEMM tile configurations that produce the session's largest single gain (+17.3\%). Each change alters the execution profile, and optimizations that would have been invisible early in the session become the dominant opportunity later. Failures carry signal as well: the same K-cache gather optimization that produced gains in three independent sessions reverted in a fourth ($-$3\%), because earlier changes in that session had altered the memory layout it depends on. An agent that cannot accumulate and reason over this evidence across iterations is unable to exploit it. We dive deeper into some of these case studies in Section~\ref{sec:exp-inference}.

\subsection{Agent State: Optimization Tree and Knowledge Ledger}
\label{sec:agent-state}

To give the agent persistent memory across iterations, Arbor maintains two inspectable data structures that are presented to the agent at the start of every iteration, independent of LLM context length. Both are shared across all policies: the agent-driven baseline reads and writes to the same tree and ledger as the structured policies.

The \emph{optimization tree} organizes the search space into category nodes (e.g., ``Sparse Attention,'' estimated at 30\% of the optimization opportunity) containing leaf nodes that each describe a concrete target. Category values are estimated during an initial profiling probe in which the agent reads the codebase, profiles the running system, and proposes targets without making code changes. Each leaf tracks its status (untried, kept, reverted, or crashed) and the throughput delta it produced. After every outcome the harness updates the tree: a KEEP raises the category's empirical success rate, a REVERT or crash lowers it. How the harness uses these scores depends on the search policy: the bandit re-ranks categories via UCB1, the sweep decides when to deepen versus advance, and the swarm dispatches workers to the highest-value untried targets. In each case the harness, not the agent, makes the traversal decision based on accumulated evidence. The agent-driven baseline reads the same tree but selects its own targets; in practice it skims across categories rather than committing to the difficult multi-iteration work that harness-driven policies enforce (Section 4). Because the evidence is explicit and inspectable rather than implicit in the LLM's context window, the agent does not forget prior outcomes or overlook unexplored categories regardless of which policy drives the search. The tree also expands during the campaign: the agent can propose new leaves or entirely new categories discovered through profiling during any iteration, so successful optimizations that shift the bottleneck to previously invisible components generate new branches for future exploration.

The \emph{knowledge ledger} complements the tree by tracking what was learned rather than what was tried. It records measured facts that are automatically flagged when subsequent changes invalidate their assumptions, unverified assumptions each carrying the cheapest experiment that would disprove them, named unknowns blocking downstream decisions, and dead ends with explicit conditions for re-opening. For example, in one session the agent discovers during profiling that a fast quantization dispatch path exists but is never activated because the kernel dispatch table lacks tuned configurations for the model's specific matrix shapes. The agent records this as an UNKNOWN with a concrete falsifier (check whether the missing entries can be added), noting that three downstream leaves are blocked on it. Resolving this UNKNOWN produces the session's largest single gain (+17.3\%). Open uncertainties are ranked by \emph{probe value}, a priority that reflects how much downstream work an uncertainty blocks per unit cost to resolve, giving the agent a concrete ranking of what to investigate next.

\subsection{End-to-End Validation}
\label{sec:validation}

Arbor gates every code-changing iteration through end-to-end validation regardless of which search policy selected the candidate. Each agent operates in an isolated workspace (a containerized git worktree), so the harness has full control over what code is deployed, can roll back any change atomically, and can verify that the benchmark and correctness scripts, specified by the task definition and read-only to the agent, have not been modified (ensuring integrity).

Each iteration produces one of three outcomes. A \emph{KEEP} indicates the change cleared both the benchmark bar and the correctness threshold; it becomes the new baseline for subsequent iterations. A \emph{REVERT} indicates the change failed validation; the workspace rolls back to the previous verified state and the category's empirical success rate in the tree decreases. A \emph{NOT\_MEASURED} indicates that the benchmark did not complete due to infrastructure failure; the change is preserved and the category's score is not penalized, so transient infrastructure issues do not bias the search.

The harness also encourages the agent to invest time in understanding before committing to changes. When the agent is uncertain about a bottleneck's root cause, unsure which code path is active, or unable to estimate the potential gain, it can declare a \emph{learn} iteration: profiling, reading code, running diagnostics, and growing the optimization tree with new candidates, all without being penalized for not producing a performance improvement. Without this distinction, the agent is pushed to skip investigation and rush to untargeted code changes.

\subsection{Search Policies}
\label{sec:policies}

Different optimization landscapes reward different traversal strategies, and a key empirical finding of this work is that the choice of policy influences the kind of optimizations the agent produces (Section~\ref{sec:experiments}). Arbor implements three structured policies and one agent-driven baseline. All four share the same agent (\texttt{Claude-Opus-4.6}), execution infrastructure, validation layer, task definition, optimization tree, knowledge ledger, and hardware. The only difference is target selection: in the structured policies, the harness selects the target and assigns the agent a specific optimization to implement; in the agent-driven baseline, the agent selects its own targets each iteration.

\paragraph{Sweep: depth-first traversal.} The sweep policy traverses categories in the order established by profiling, advancing to the next only when all leaves in the current category are exhausted. Within a category, KEEPs spawn depth-expansion children, enabling the agent to refine a successful optimization across multiple iterations (prototyping, debugging, tuning) before the harness moves it on. This sustained focus can produce kernel-level rewrites that broader policies do not reach (Section~\ref{sec:experiments}).

\paragraph{Bandit: outcome-aware re-ranking.} The bandit policy extends the sweep's category structure with adaptive selection. Rather than following a fixed order, the harness scores categories with a UCB1-derived criterion that balances estimated value against observed outcomes:
\begin{equation}
\label{eq:ucb1}
\text{score}(c) = \hat{v}_c \cdot r_c \;+\; C\sqrt{\tfrac{\ln N}{n_c}}
\end{equation}
where $\hat{v}_c$ is the estimated value of category $c$ (from profiling), $r_c$ is the empirical success rate (keeps divided by trials, initialized to 0.5 for untried categories), $N$ is the total iterations completed, $n_c$ is attempts in category $c$, and $C = 1.41$. After each outcome the harness re-ranks: when failures lower a category's $r_c$, untried categories with prior 0.5 rise above it. On DSv4.1-Flash, UCB1 redirects from Sparse Attention (estimated at 30\%) to mHC Fusion (estimated at 10\%) after failures accumulate, where the agent writes a fused Triton kernel eliminating 80~kernel invocations per forward pass (+19.4\%). Replaying the trajectory under $C \in [0, 2.82]$ confirms that all values select the same winning category at the same step; re-ranking after failures, not the exploration constant, is the load-bearing mechanism. We dive deeper into the optimization landscape coverage achieved by different policies in a fixed time period in Section~\ref{sec:exp-inference}.

\paragraph{Swarm: parallel dispatch.} The swarm policy dispatches $N$ workers to $N$ distinct optimization targets in parallel, each constrained to disjoint source files. A selector agent reviews the combined diff and the harness validates the merged result end-to-end as a single unit, amortizing one expensive benchmark across several changes. The tradeoff is token cost: $N$ parallel workers consume roughly $N\!\times$ the tokens of a single-agent iteration.

\paragraph{Agent-driven baseline.} The agent-driven baseline imposes no harness-enforced traversal order. The agent reads the same optimization tree and knowledge ledger as the structured policies and has the same \emph{learn} budget, but decides each iteration what to investigate, what to implement, and when to move on. The same validation layer and execution infrastructure apply. This is the critical comparison: the agent-driven baseline receives every piece of infrastructure that makes sustained optimization possible: containerized isolation, automatic rollback, end-to-end validation, persistent state, the optimization tree, and the knowledge ledger. The only thing it lacks is harness-driven target selection.

\section{Evaluation Framework}
\label{sec:eval}

The system design in Section~\ref{sec:design} introduces multiple search policies that produce different outcomes from the same agent, tools, and hardware. Final performance improvement alone cannot distinguish them. We want to understand the trajectory the agent took to get there: did it persist through difficult optimizations or give up after the first failure? Did it produce large, targeted code changes or many small speculative edits? Did it explore the full optimization space, or get stuck in one corner? These questions require metrics that capture agent behavior across the full optimization trajectory, not just the endpoint.

We organize metrics into four dimensions: \emph{efficiency} (how much of the compute budget translates into improvement and how decisively the agent uses it), \emph{grounding} (whether the agent's measurements are trustworthy), \emph{robustness} (how the agent handles failures and dead ends), and \emph{depth} (what the agent built and how much of the space it covered). All metrics are computable from session artifacts without additional instrumentation, and the framework applies to any agentic system running write-evaluate-improve loops. Table~\ref{tab:metrics} lists the metrics and their definitions; full numerical values across all four policies appear in the appendix (Table~\ref{tab:eval-full}).

\begin{table}[t]
\caption{Evaluation framework: four dimensions and their metrics. All metrics are computed from session artifacts: iteration logs, token counts, the optimization tree, and agent transcripts.}
\label{tab:metrics}
\vskip 0.15in
\begin{center}
\begin{small}
\begin{tabular}{@{}llp{3.0cm}@{}}
\toprule
Dimension & Metric & Definition \\
\midrule
\multirow{4}{*}{Efficiency}
 & Gain / token cost & $\Delta$\% $\div$ relative token cost \\
 & Gain / GPU-hr & $\Delta$\% $\div$ (wall-hrs $\times$ GPUs) \\
 & Gain / iteration & $\Delta$\% $\div$ completed iterations \\
 & Decisiveness & 1 $-$ reversals $\div$ reasoning blocks \\
\midrule
\multirow{2}{*}{Grounding}
 & Integrity & benchmarked $\wedge$ files unmodified \\
 & Research rate & researched $\div$ total edits \\
\midrule
\multirow{2}{*}{Robustness}
 & Recovery rate & recovered $\div$ failures \\
 & Evidence on abandon & investigated $\div$ abandonments \\
\midrule
\multirow{2}{*}{Depth}
 & Keep rate & KEEPs $\div$ total attempts \\
 & Coverage & categories visited $\div$ declared \\
\bottomrule
\end{tabular}
\end{small}
\end{center}
\vskip -0.1in
\end{table}

\paragraph{Efficiency: gain per unit of budget.} Four metrics capture how much of the agent's compute budget translates into actual improvement. Three normalize gain by a different resource: \emph{gain per token cost} divides the total percentage-point improvement by the relative token expenditure of the run, so that a swarm of three parallel workers that consumes $3\times$ the tokens of a single-agent policy is penalized accordingly; \emph{gain per GPU-hour} divides improvement by wall-clock hours multiplied by the number of GPUs, measuring how efficiently the policy converts hardware time into throughput; and \emph{gain per iteration} divides improvement by the number of completed think-implement-validate cycles. The fourth metric, \emph{decisiveness}, measures whether the agent uses its reasoning budget to drive forward or wastes it on vacillation: for each reasoning block in the agent's transcript, we check whether the agent reverses a decision it previously committed to (e.g., abandoning an implementation mid-way to restart with a different approach). A score of~1.0 means every reasoning block drives forward; lower scores indicate the agent is spending tokens oscillating between topics rather than committing to one direction and implementing it.

\paragraph{Grounding: measurement trustworthiness.} Autonomous agents operating for hours without supervision can drift: editing evaluation scripts to inflate scores, or making code changes without first profiling the system. \emph{Integrity} is a binary check: after each iteration, the harness verifies file-digest checksums on benchmark scripts and environment configuration; a score of~1.0 means no evaluation artifact was modified. \emph{Research rate} is the fraction of implementation iterations where the agent performed at least one investigative action (reading code, running a profiler, or examining prior results) before writing its first edit. A research rate below~1.0 indicates the agent is writing speculative code without grounding it in the system's current state. All four policies score~1.0 on both metrics, because Arbor delegates benchmarking, correctness checking, and code bookkeeping to the harness, removing the opportunity for drift regardless of search policy.

\paragraph{Robustness: persistence through failure.} Long-running campaigns inevitably encounter failures: infrastructure crashes, regressions, and dead ends. \emph{Recovery rate} is the fraction of failures after which the agent resumes productive work within the next iteration, rather than entering a failure spiral. \emph{Evidence on abandonment} checks whether the agent performed diagnostic work before moving on from a failed target, producing reusable signal in the knowledge ledger rather than silently discarding information. Recovery rates range 0.88--0.92 across policies; the harness enforces recovery by reverting failed changes and reassigning targets.

\paragraph{Depth: what the agent built.} Two metrics capture the character of the optimizations the agent produces. \emph{Keep rate} is the fraction of implementation attempts that clear both the benchmark bar and the correctness gate and become the new baseline; it is the yield of the agent's implementation effort. \emph{Coverage} is the fraction of the declared optimization categories that the policy visits at least once. A sweep that concentrates in a single category scores~1/5; a swarm that dispatches workers across all categories scores~5/5. Coverage and keep rate together reveal whether a policy explores broadly with shallow attempts or narrows deeply with high-yield implementations.

Grounding and robustness are harness-enforced; the six metrics that differentiate policies (efficiency and depth) are visualized in Section~\ref{sec:experiments} (Figure~\ref{fig:radar}).

\section{Experiments}
\label{sec:experiments}

We first demonstrate Arbor's effectiveness on production workloads across domains and hardware (\S\ref{sec:exp-breadth}), then conduct a controlled policy ablation on DSv4.1-Flash analyzing how search policy shapes agent behavior, the optimizations discovered, and code quality (\S\ref{sec:exp-inference}), and finally demonstrate generality on training workloads (\S\ref{sec:exp-training}). In every experiment, the execution infrastructure and validation layer are identical; only the task definition and, in the controlled ablation, the search policy varies.

\subsection{Arbor at Scale}
\label{sec:exp-breadth}

\begin{figure}[t]
\centering
\includegraphics[width=\linewidth]{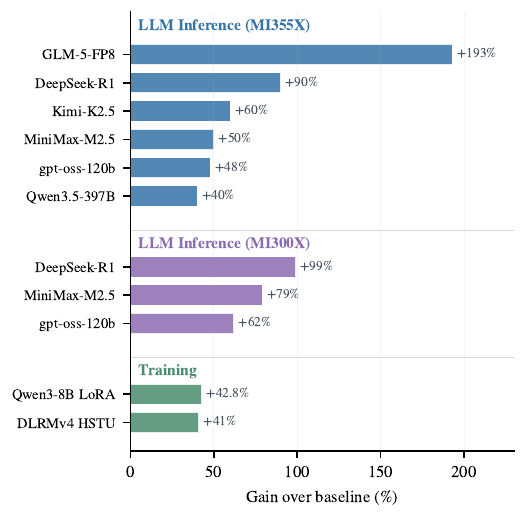}
\caption{Arbor results across domains and hardware. LLM inference 
(blue): +40--193\% throughput-interactivity Pareto improvement over 
InferenceX vendor baselines on six production models (MI355X) and 
three models (MI300X), gated on throughput, latency, and accuracy. 
Training (green): +42.8\% on Qwen3-8B LoRA fine-tuning, +41\% 
cumulative on DLRMv4 recommendation training.}

\label{fig:breadth-bars}
\end{figure}

Figure~\ref{fig:breadth-bars} summarizes Arbor's results across three domains and two GPU generations. On LLM inference, Arbor achieves +40\% to +193\% throughput-interactivity Pareto improvement over vendor-optimized InferenceX baselines across six production models on MI355X, gated on throughput, latency, and accuracy; the gains hold across the full concurrency range, not only at a single operating point (full Pareto curves in the appendix (attached as supplementary material), Figure~\ref{fig:pareto-mi355x}). Three of these models are also evaluated on MI300X, where gains range from +62\% to +99\% with no modification to the system; the profiling-driven search adapts to the different bottleneck distribution automatically. The largest gain (+193\% on GLM-5-FP8) comes from co-optimizing tensor parallelism configuration alongside attention kernels and MoE dispatch, a compound intervention spanning multiple stack layers. Each breadth result is the best of multiple campaigns on the respective model; the controlled ablation in \S\ref{sec:exp-inference} reports variance across four independent runs for all the Arbor search policies individually.

On LLM training, the swarm policy achieves +42.8\% throughput on Qwen3-8B LoRA fine-tuning (8$\times$~MI355X). On recommendation model training, Arbor produces a cumulative +41\% gain on DLRMv4 (MI350X), with agents autonomously writing and validating kernel-level and framework-level optimizations spanning thousands of lines of code, including sync-free gradient accumulation (+4.8--8.5\%), fused layer-norm and embedding operations (+3.96\%), and FP8 GEMM chain wiring (+3.59\%), all verified convergence-safe against MLPerf RCP floors.
The same execution infrastructure runs unchanged across all three domains; only the task definition changes.
We dive deeper into some of these results, the search policy ablations, and the evaluation framework in the next sections. 

\subsection{Policy Comparison: DSv4.1-Flash Inference}
\label{sec:exp-inference}

To understand how search policy shapes agent behavior, we conduct a controlled ablation on the AgentX workload over DSv4.1-Flash, holding the agent, tools, hardware, and validation constant while varying only the search policy.

\paragraph{Setup.} We evaluate on DeepSeek-V4.1-Flash with FP4 
quantization and DSpark speculative decoding, served on 
4$\times$~MI355X at tensor parallelism~4. The validation benchmark 
is a 600\,s AgentX replay at concurrency~32, a proxy of the full 
one-hour trace that preserves the same traffic shape at lower cost. 
Gains measured on the proxy transfer to the full benchmark; all 
production results in Section~\ref{sec:exp-breadth} are confirmed on 
the full evaluation. All four policies use the same agent 
(\texttt{Claude-Opus-4.6}), the same execution infrastructure, and 
the same correctness gate, each starting from a fresh container.

\begin{figure}[t]
\centering
\includegraphics[width=\linewidth]{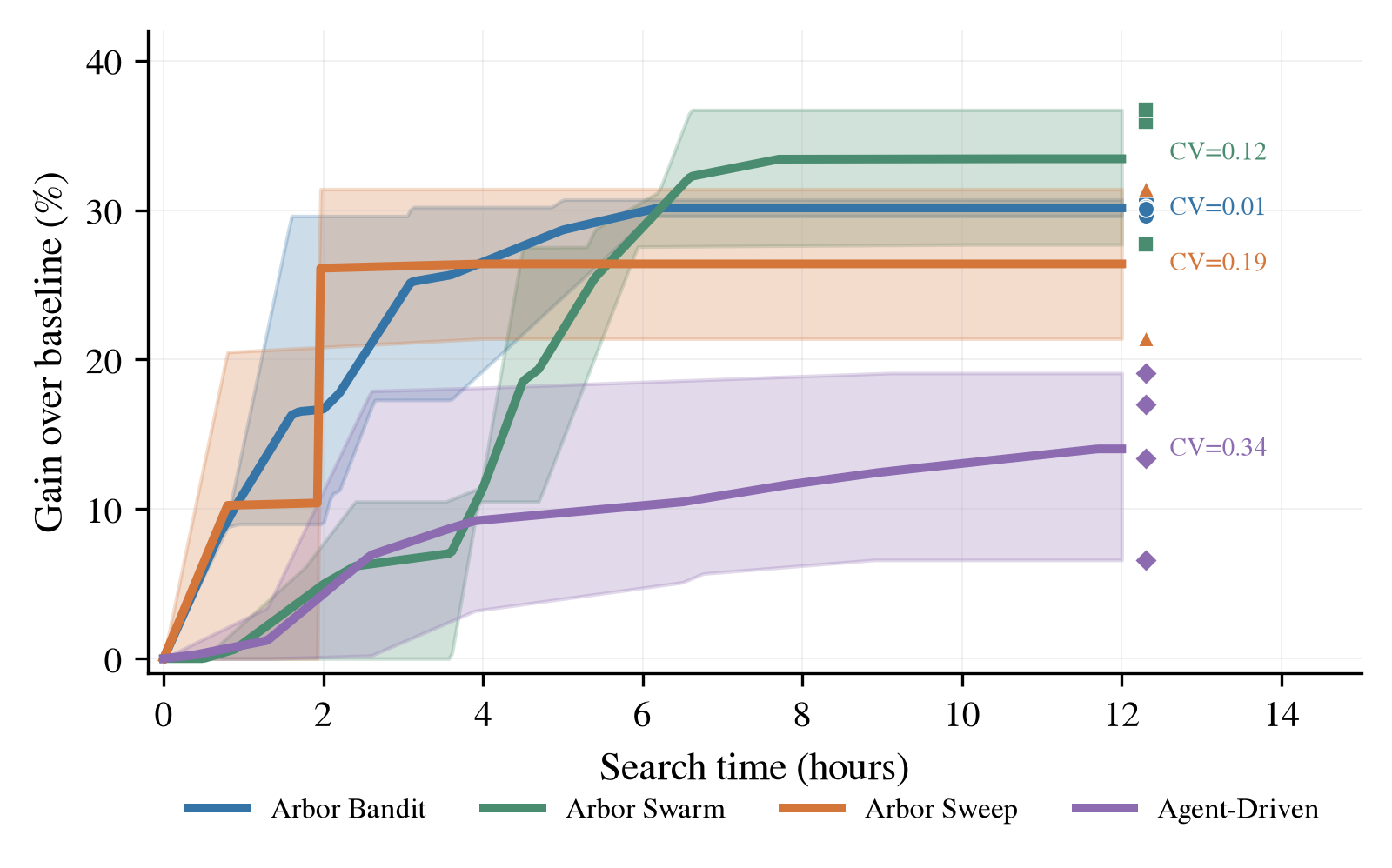}
\caption{Best-so-far throughput gain on DSv4.1-Flash inference 
across four independent runs per policy (12\,h budget). Bold lines: 
mean across runs. Shaded bands: min--max range. Dots: individual 
run endpoints. The bandit converges to +30.1\% (CV$=$0.01) within 
6\,h; the agent-driven baseline reaches +14.0\% (CV$=$0.34) with 
gains ranging from +6.6\% to +19.1\%.}
\label{fig:gain-curves}
\end{figure}

\paragraph{Results.} Figure~\ref{fig:gain-curves} shows the best-so-far gain curves across four runs per policy with min--max variance bands.

Every structured policy outperforms the unstructured baseline (agent-driven) at matched budget. The bandit reaches +30.1\% with four runs landing within a 1.1~percentage-point range (coefficient of variation CV$=$0.01), converging well before the 12\,h budget expires. The swarm achieves the highest absolute gain (+33.4\%) and the highest per-iteration efficiency (4.08\% gain/iter), producing a staircase of compounding optimizations where each KEEP resolves a prerequisite for the next. The sweep reaches +26.4\% through deeper exploration within each category it visits, finding different winning components than the bandit. The agent-driven baseline at 12\,h reaches +14.0\% mean with the widest variance (CV$=$0.34): across four runs, gains range from +6.6\% to +19.1\%, and the agent follows a different path each time.

We also run agent-driven sessions beyond the 12\,h budget to measure convergence. All four runs eventually reach the bandit's +30\%, but the time to do so ranges from 16 to 30~hours (individual trajectories in Figure~\ref{fig:extended}), requiring 1.3--2.5$\times$ the wall-clock time of the structured policies with high inter-run variance. 

\begin{figure*}[t]
\centering
\includegraphics[width=\textwidth]{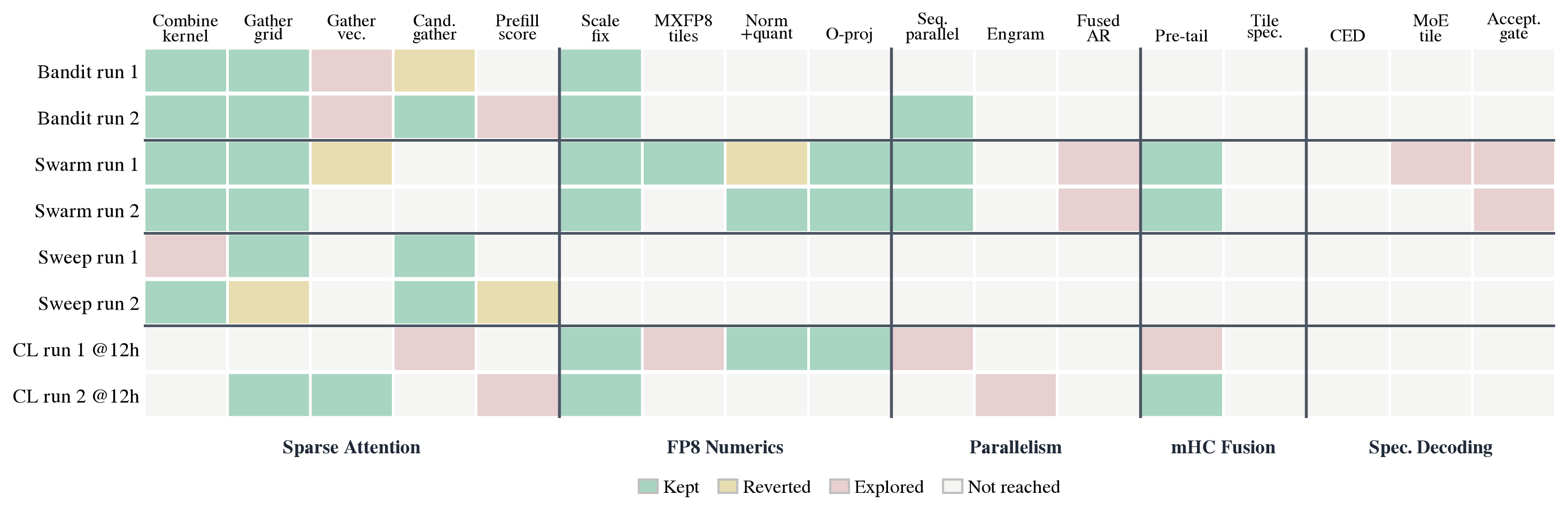}
\caption{Optimization tree coverage for DSv4.1-Flash (best two runs 
per policy). Columns: 17 leaf targets across 5 profiled categories. 
Green: kept (cleared benchmark and correctness). Yellow: reverted. 
Pink: explored but not implemented. White: not reached. The sweep 
concentrates in Sparse Attention with multiple depth-expansion KEEPs 
per target; the bandit and swarm spread across categories; the 
agent-driven baseline covers broadly but shallowly.}
\label{fig:coverage}
\end{figure*}

\paragraph{Coverage and depth.} Figure~\ref{fig:coverage} shows leaf-node outcomes for two representative runs of each policy. The bandit and swarm explore targets across most categories; the sweep concentrates entirely within Sparse Attention but goes significantly deeper within it. On the K-cache gather grid optimization (``Gather grid'' in Figure~\ref{fig:coverage}), the bandit applies a compute unit derived worker-count formula in a single iteration and moves on. The sweep refines this same target across three consecutive KEEPs: first the same basic fix, then extending it to the DSv4 codebase (which shares the kernel but uses a different call site), then replacing the simple formula with a more hardware-informed variant that accounts for GPU occupancy targets and CU-level wave scheduling. The bandit and swarm, facing many categories competing for attention, do not spend enough consecutive iterations in one area to reach this level of depth. However, they cover more topics and improve the state of the inference stack more holistically. In the agent-driven baseline, despite the high coverage in Figure~\ref{fig:coverage}, the agent skims through most of the topics without sustained focus, leading to the lower performance gain than structured search policies as shown in Figure~\ref{fig:gain-curves}. 

\paragraph{Implementation variance across runs.} Even within the same policy and the same tree assignment, the agent's implementation varies across runs.

Both bandit runs attempt the FP8 scale-layout fix (``Scale fix'' in Figure~\ref{fig:coverage}) at iteration~2, receiving the same tree state and profiling context. Run~1's agent identifies two issues in a single iteration: the scale-layout detection bug and a missing pre-quantized bypass in the projection dispatch. It adds an override that routes already-quantized FP8 data directly to the GEMM, producing +18.6\% in one step. Run~2's agent fixes only the detection logic, which alone produces +0.9\% because the fused dispatch path does not activate under this model's quantization group shape. Run~2 discovers this gap five iterations later during a profiling probe, but has moved to other categories by then.

The ``Cand.\ gather'' column in Figure~\ref{fig:coverage} shows the same pattern: green/KEEP in one run, yellow/REVERT in another. Both agents build the same Triton kernel that compacts ${\sim}$16K candidate columns from a 1M-wide logit buffer. Run~1 integrates the kernel into the MLA attention path; run~2 integrates it into the sparse indexer path with explicit pointer-stable buffers for CUDA-graph replay. Run~1's integration breaks graph capture in the tensor-parallel server and reverts ($-$36\%); run~2's is kept (+11\%). The optimization target and kernel logic are identical however, the integration decision and agent thinking determined the end result.

The tree structure ensures that both agents reach the same targets at similar iterations. The agent's implementation decisions like where to integrate a kernel, whether to bundle additional code paths, whether to ensure graph-capture compatibility, are the source of per-iteration variation. Despite them, the final gains converge: run~1 reaches +29.6\% through 3~KEEPs, run~2 through 5~KEEPs across broader coverage.

\begin{figure}[t]
\centering
\includegraphics[width=0.85\linewidth]{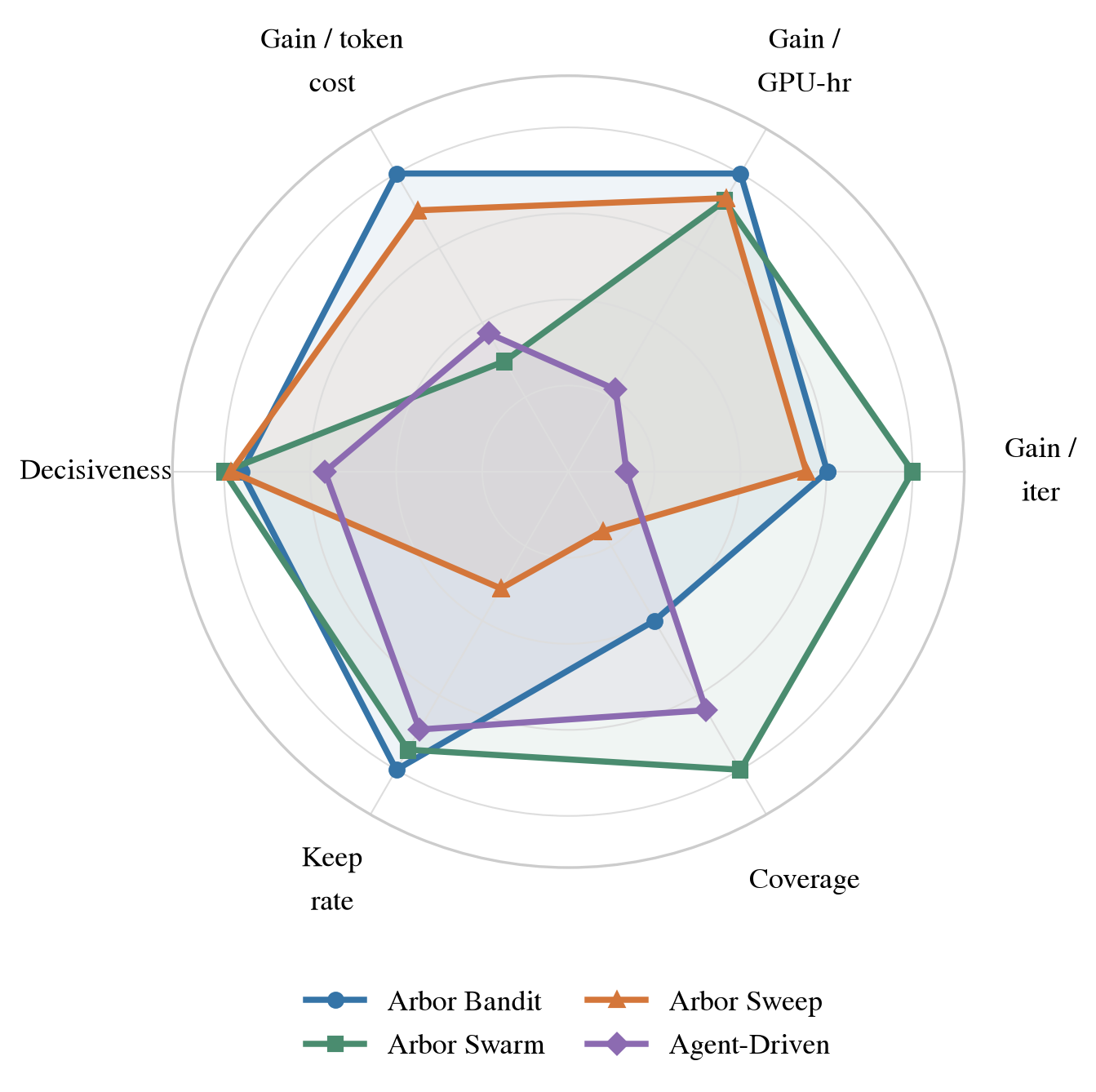}
\caption{Radar chart of the six differentiating evaluation metrics 
from Section~\ref{sec:eval} (mean across runs per policy). Each 
polygon is one policy. The bandit (blue) is the most balanced; the 
swarm (green) dominates on gain per iteration and coverage; the sweep (orange) collapses on coverage but achieves strong gain per GPU-hour; 
the agent-driven baseline (purple) scores lowest on efficiency 
metrics. Full numerical values in Table~\ref{tab:eval-full}.}
\label{fig:radar}
\end{figure}

\paragraph{Evaluation framework.} Figure~\ref{fig:radar} applies the six differentiating metrics from Section~\ref{sec:eval} to all four policies (full numerical values in the appendix, Table~\ref{tab:eval-full}). Each policy occupies a distinct region of the metric space. The \emph{bandit} has the largest and most balanced polygon: highest gain per GPU-hour (1.09\%), highest keep rate (0.74), high decisiveness (0.94), and the highest gain per token cost (0.50), reflecting targeted changes that survive validation at low token expenditure. The \emph{swarm} dominates on gain per iteration (4.08\%) and coverage (5/5), compounding parallel changes across all categories, but its $3\times$ token cost from parallel workers pulls gain per token cost to the lowest (0.19); its decisiveness is the highest of any policy (0.99) because each worker commits to a single assigned target. The \emph{sweep} achieves strong gain per GPU-hour (1.00\%) and high decisiveness (0.97) but collapses on coverage (1/5) and keep rate (0.29), spending its entire budget in Sparse Attention with many reverts along the way to its deep kernel rewrites. The \emph{agent-driven baseline} scores lowest on gain per iteration (0.69\%) and gain per GPU-hour (0.30\%), with low gain per token cost (0.23) and the lowest decisiveness (0.77); its coverage (4/5) reflects broad but shallow exploration.

Structure improves not only the magnitude of the gain but the efficiency of the code the agent writes. At matched 12\,h budget, the bandit produces ${\sim}$400~lines of added code for +30.1\% gain (13~LOC per percentage point), while the agent-driven baseline produces ${\sim}$850~lines for +14\% (61~LOC per percentage point), a ${\sim}$4$\times$ difference. The bandit's code is concentrated in a few high-impact modifications whereas the agent-driven baseline writes more speculative code across a wider set of files.

The decisiveness gap (bandit 0.94 vs.\ agent-driven 0.77) explains this difference. Counting direction-change phrases in agent transcripts (``let me try a different approach,'' ``let me step back,'' ``this is getting complex,'' ``I'm going in circles''), the agent-driven baseline produces ${\sim}$13 per session compared with ${\sim}$5.5 for the bandit, ${\sim}$6 for the swarm, and ${\sim}$4 for the sweep. Critically, the \emph{nature} of these changes differs. The structured policies' direction changes are implementation-level: retrying a failed docker copy, adjusting a Triton tile size, or switching between equivalent code paths to fix a build error, all tactical debugging within a harness-assigned target, none of which change what the agent is optimizing. The agent-driven baseline's direction changes are strategic: the agent starts investigating one optimization target, encounters complexity or ambiguity, and pivots to a different part of the stack entirely. In one session the agent writes ``I've been fighting this crash for hours \ldots let me take a step back'' before abandoning the approach, and in another ``I'm going around in circles \ldots let me try a completely different approach,'' switching from a memory-layout fix to an unrelated scheduling change mid-iteration. 

\paragraph{Why end-to-end validation matters.} Across our campaigns, kernel-level improvements that pass local correctness and speed gates frequently regress end-to-end throughput. Three representative cases illustrate distinct failure modes.

\emph{Operating-point mismatch.} An MQA logits kernel rewrite narrows the computation width, skipping columns that do not contribute to top-K selection. The rewrite runs 2$\times$ faster locally and produces identical outputs. Under production traffic, throughput regresses by 38\%: the narrowed output changes the input shape to the downstream top-K stage, which was tuned for the wider layout, and a single long-context request in the batch (which sets \texttt{max(seq\_lens)} for all rows) forces the slow path for the entire batch. No component benchmark reproduces this because it requires the production request-length distribution.

\emph{Amdahl ceiling with closed-loop feedback.} A kernel optimization delivers 1.26$\times$ local speedup on a component accounting for 13.5\% of execution time but produces only +1.5\% end-to-end gain, less than half what Amdahl's Law predicts. The reason is that the serving framework's closed-loop scheduler reduces the in-flight batch size in response to faster decoding, restoring per-token cost in all batch-size-dependent components. A component benchmark shows +26\%; the full system delivers +1.5\%.

\emph{Compounding interaction.} A standard distributed-training flag (overlapping parameter gathering with compute) improves throughput by 5\% on one configuration but regresses it by 35\% on another where a fused cross-entropy kernel is active. The overlap and the fusion compete for memory bandwidth during the backward pass. The flag is a well-tested framework feature; the regression arises only from the specific combination of optimizations active in the stack.

All three cases are invisible to component-level benchmarks and would corrupt the search if accepted without end-to-end validation.

\subsection{Training: LLM and Recommendation}
\label{sec:exp-training}

To demonstrate generality beyond inference, we apply Arbor to two training workloads using the same execution infrastructure with only the task definition changed.

\paragraph{Qwen3-8B LoRA.} On Qwen3-8B LoRA fine-tuning (rank$=$16, BF16, FSDP2, 8$\times$~MI355X), Arbor achieves +42.8\% training throughput improvement through DDP flag tuning, torch.addmm fast-path wiring, and in-place LoRA operations that eliminate per-layer allocation. As on inference, the sweep policy produces the deepest optimizations: a fused Triton cross-entropy kernel rewrite (+15.2\%) found through sustained depth in the kernel fusion category, consistent with the pattern that depth-first traversal reaches hardware-aware kernel engineering that broader policies do not. The DLRMv4 deployment timeline, showing how 14 parallel campaigns compound thousands of lines of kernel and framework code into a cumulative +41\% gain over 300 hours, appears in the appendix (Figure~\ref{fig:dlrm-timeline}).

\subsection{Analysis: How to Select a Policy}
\label{sec:exp-summary}

The experiments reveal that the best policy depends on the available budget and the desired tradeoff between depth, speed, and reproducibility. The \textbf{sweep} produces the deepest optimizations, reaching hardware-aware kernel engineering that other policies do not attempt, but requires the most wall-clock time and covers the fewest categories. The \textbf{swarm} produces the fastest gains by compounding parallel changes, but at the highest token cost and with optimizations concentrated at the config/framework level. The \textbf{bandit} balances coverage, efficiency, and reproducibility (CV$=$0.01), making it the default for budget-constrained settings where predictability matters. The \textbf{agent-driven baseline} eventually matches the bandit's gain given 1.3--2.5$\times$ the wall-clock budget, but the path is unpredictable (CV$=$0.34). The goal of this controlled ablation is to characterize how different policies push the agent to write different kinds of code and navigate the optimization space in different ways, not to declare a single winner. 
\section{Related Work}
\label{sec:related}

\paragraph{LLM-guided code optimization.} FunSearch~\cite{romeraparedes2024funsearch} introduced LLM-guided evolutionary search over short programs. AlphaEvolve~\cite{alphaevolve2025} extends this to full codebases, discovering improved algorithms for matrix multiplication and data center scheduling. AVO~\cite{chen2026avo} replaces prescribed mutation operators with an autonomous agent loop, producing attention kernels that exceed cuDNN throughput. DuoDiscover~\cite{duodiscover2026} jointly discovers optimization targets and cheap proxy evaluators, scoring candidates 22$\times$ faster than true evaluation on repository benchmarks. PerfAgent~\cite{perfagent2026} uses profiler-guided iterative refinement for repository-level optimization. Each of these systems operates within a single optimization layer against a self-contained evaluator. Arbor addresses a setting where the search space spans hundreds of interacting candidates across multiple stack layers, the objective is multi-dimensional, and successful optimizations shift the bottleneck to previously invisible components.

\paragraph{Agent-driven system optimization.} PROMPTS~\cite{prompts2026} is the closest prior work: a multi-agent planning framework for LLM training and serving optimization, achieving up to 434\% improvement across production workloads. Unlike Arbor, PROMPTS focuses on configuration and deployment planning rather than code-level optimization with iterative search over a tree of scored candidates. InferenceBench~\cite{inferencebench2026} benchmarks agents on open-ended inference optimization and finds that the bottleneck is not domain knowledge but systematic evaluation and diverse exploration, an observation consistent with Arbor's finding that structure improves code efficiency (Section~\ref{sec:experiments}).

\paragraph{Multi-agent kernel optimization.} KernelSkill~\cite{kernelskill202}, Astra~\cite{astra2026}, AccelOpt~\cite{zhang2026accelopt}, STARK~\cite{dong2025stark}, and KernelArc~\cite{kernelarc2026} introduce strategy-specialized agents and shared optimization memory for GPU kernel synthesis. These systems demonstrate that multi-agent coordination improves single-kernel outcomes but all operate within a single layer: a kernel that passes local microbenchmarks may regress end-to-end throughput due to graph capture conflicts or resource exhaustion that manifest only under full serving load (Section~\ref{sec:validation}). Arbor's validation layer gates every change through an end-to-end benchmark, preventing such regressions from corrupting the search.

\paragraph{LLM-driven tree search.} Tree of Thoughts~\cite{yao2023tree}, LATS~\cite{lats2024}, SWE-Search~\cite{swe-search2024}, AIDE~\cite{aide2025}, and MARS~\cite{mars2026} apply LLM-guided tree search to reasoning, software engineering, and ML experiment design. Concurrently, Jin et al.~\cite{jin2026arbormsr} present a system also named Arbor for autonomous scientific research using hypothesis-tree refinement. Our work applies scored traversal to production software stack optimization, where each evaluation runs an end-to-end serving benchmark, outcomes are noisy, and the action space expands dynamically as bottlenecks shift.

\paragraph{Agent evaluation.} Li et al.~\cite{harnessbench2026} demonstrate that the same model's ranking can reverse by over 30 points depending on the harness, establishing that the harness matters as much as the agent. TRACE~\cite{trace2026} proposes trajectory-aware evaluation of process efficiency and reasoning robustness for research agents. Arbor's evaluation framework extends this direction to long-running optimization campaigns with metrics specific to the write-evaluate-improve loop.
\section{Conclusion}
\label{sec:conclusion}

Arbor separates search policy from execution infrastructure, enabling the first controlled comparison of traversal strategies for agent-driven system optimization. The central finding is that search policy shapes not only the magnitude but the nature of the optimizations: the bandit writes targeted framework fixes with low variance (CV$=$0.01), the sweep produces deep kernel rewrites, and the swarm compounds independent changes at the highest per-iteration efficiency. All three outperform unstructured search at equal budget, 
and the infrastructure runs unchanged across inference, 
training, and recommendation workloads.

\bibliography{example_paper}
\bibliographystyle{mlsys2025}



\clearpage
\appendix
\section{Extended Agent-Driven Trajectories}
\label{app:extended}

\begin{figure}[h]
\centering
\includegraphics[width=\linewidth]{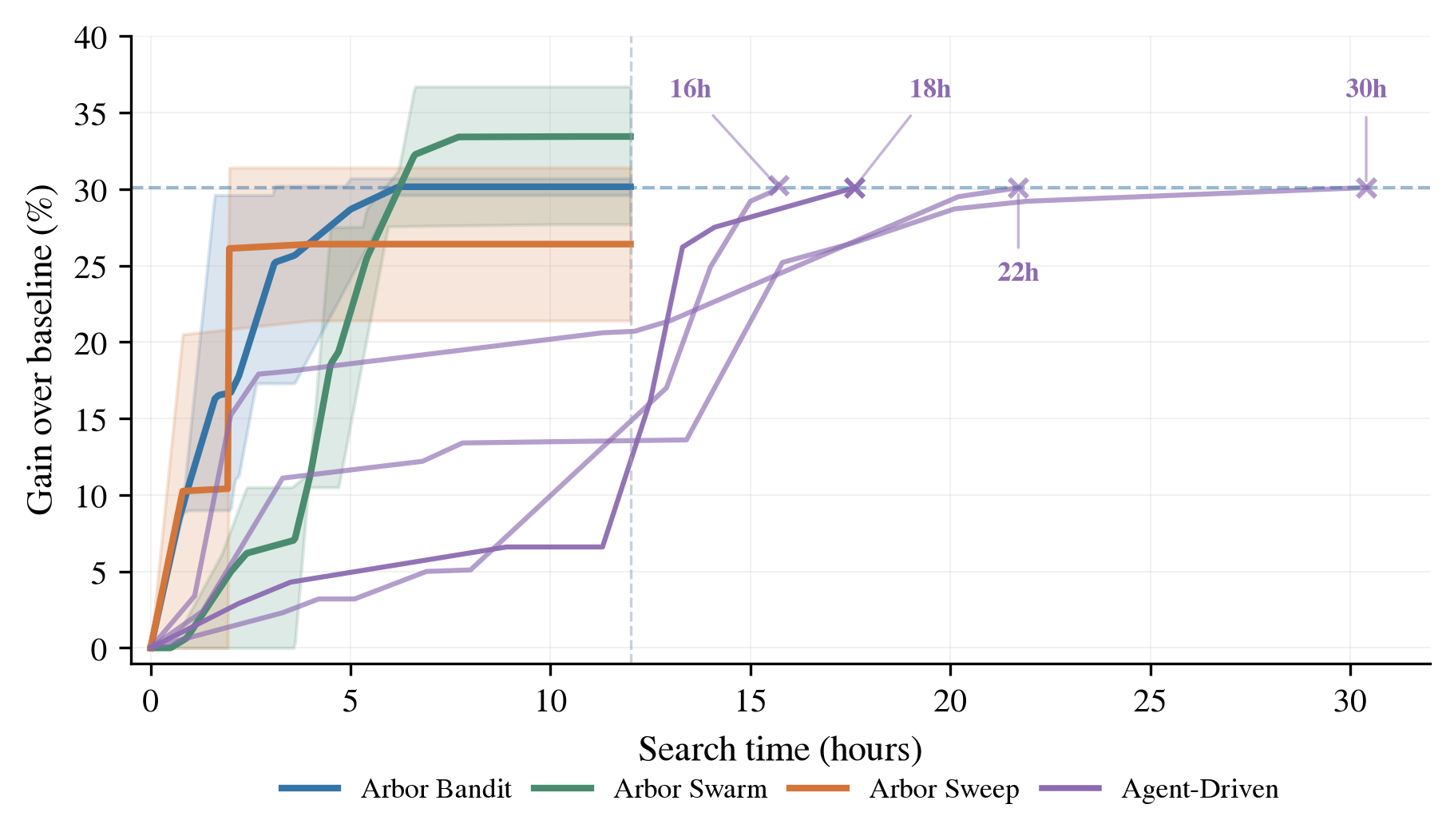}
\caption{Extended view showing individual agent-driven trajectories. Structured policies (bandit, swarm, sweep) converge by 12\,h (dashed line). The agent-driven baseline's four runs catch up at 16, 18, 22, and 30~hours respectively, with higher inter-run variance at every time point.}
\label{fig:extended}
\end{figure}

\section{DLRMv4 Deployment Timeline}
\label{app:dlrm}
\begin{figure}[h]
\centering
\includegraphics[width=\linewidth]{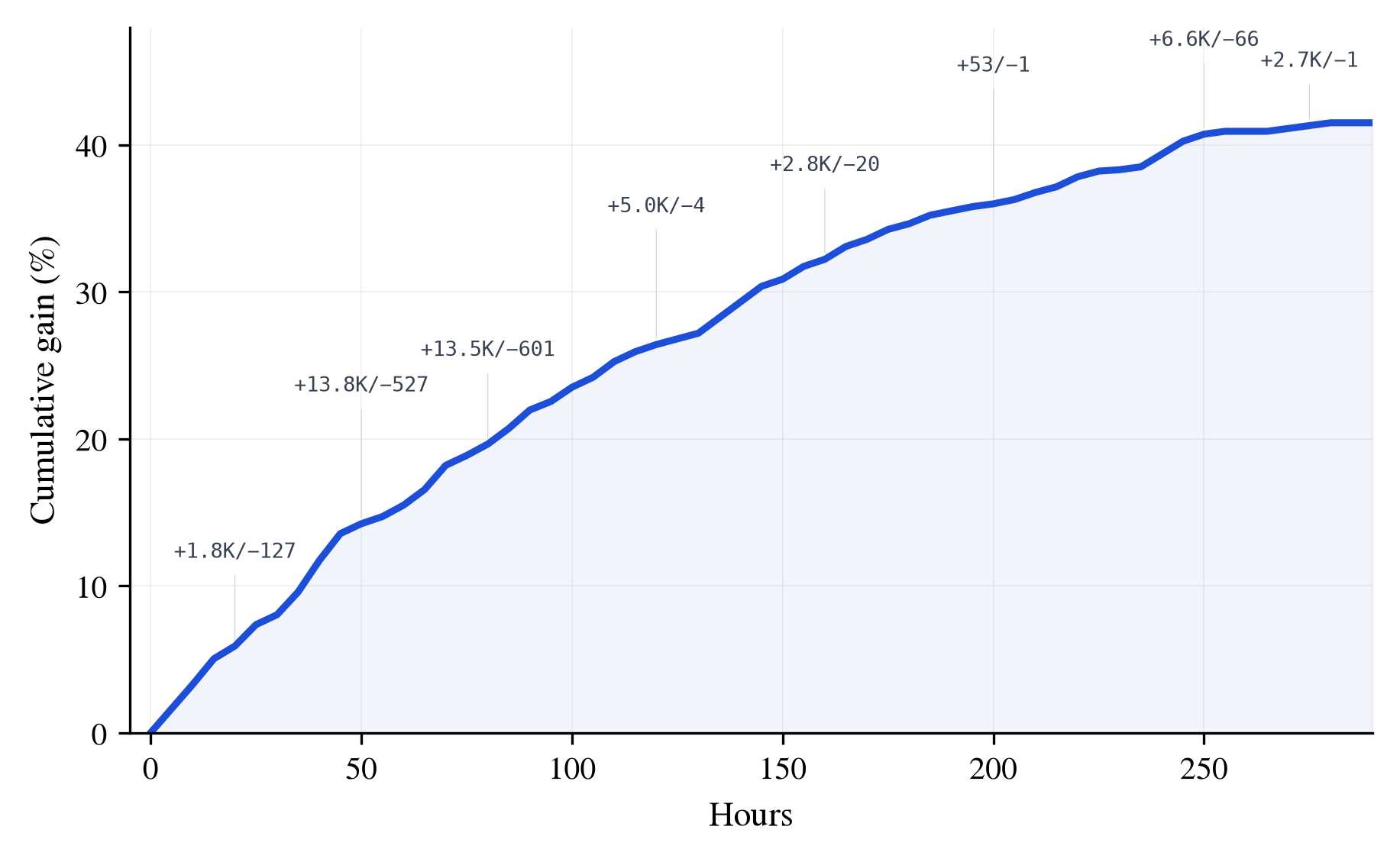}
\caption{DLRMv4 cumulative gain over 300 hours across 14 parallel campaigns. Annotations show lines of code added/removed at major ship milestones. The near-linear growth with no plateau demonstrates sustained optimization at production scale.}
\label{fig:dlrm-timeline}
\end{figure}

Figure~\ref{fig:dlrm-timeline} traces the cumulative throughput 
gain across the Arbor campaign on DLRMv4 recommendation 
model training (MI350X), spanning 832 iterations. The campaign targets different areas of the stack: 
\texttt{c8num} focuses on FP8 numerics and sync-free gradient 
accumulation, \texttt{r14} on the HSTU FP8 GEMM chain, 
\texttt{r16} on layer-norm and embedding fusion, \texttt{dlstart} 
on cold-start and input pipeline, and \texttt{comm} on multi-node 
communication. The LOC annotations mark the KEEP milestones where 
validated changes from the agent were merged. Early ships land configuration and framework-level 
fixes; the two largest ships (+13.8K/--527 and +13.5K/--601 LOC) 
integrate custom FP8 kernels and fused backward passes that 
required multi-iteration development across several campaigns. 

\section{InferenceX Pareto Curves}
\label{app:inferencex}

\begin{figure*}[h]
  \centering
  \includegraphics[width=0.32\textwidth]{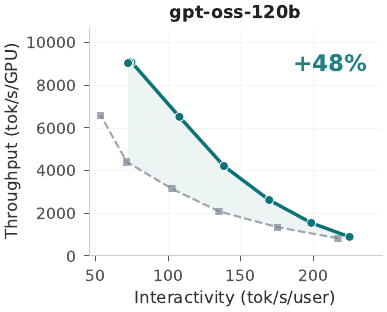}
  \hfill
  \includegraphics[width=0.32\textwidth]{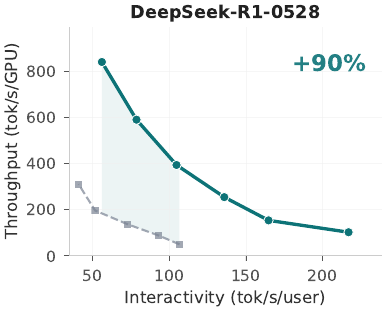}
  \hfill
  \includegraphics[width=0.32\textwidth]{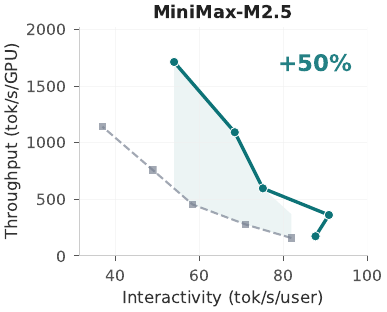}\\[4pt]
  \includegraphics[width=0.32\textwidth]{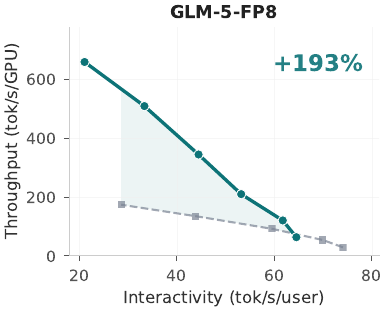}
  \hfill
  \includegraphics[width=0.32\textwidth]{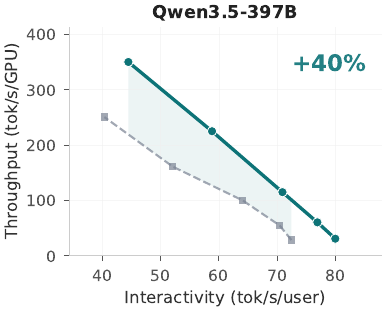}
  \hfill
  \includegraphics[width=0.32\textwidth]{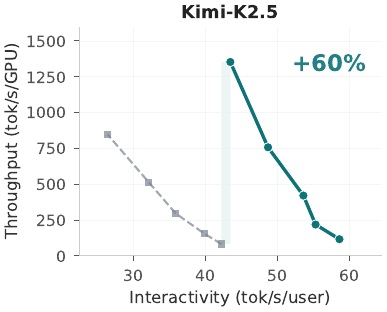}\\[4pt]
  \includegraphics[width=0.32\textwidth]{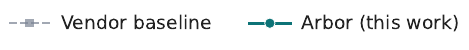}
  \caption{Throughput vs.\ interactivity ($1/\mathrm{TPOT}$) on AMD MI355X for six production models. Higher is better on both axes. Percentages indicate output throughput gain at concurrency 64 relative to the InferenceX vendor baseline.}
  \label{fig:pareto-mi355x}
\end{figure*}

Figure~\ref{fig:pareto-mi355x} plots the throughput--interactivity Pareto frontier for each of the six production models on MI355X before and after Arbor optimization. Following InferenceX~\cite{inferencex}, interactivity is defined as $1/\mathrm{TPOT}$ (tokens per second per user). The Arbor-optimized curve dominates the baseline Pareto frontier on all six models: throughput increases at every interactivity level, not only at a single operating point. All benchmarks use ISL$=$OSL$=$1024.

\begin{figure*}[h]
  \centering
  \includegraphics[width=0.32\textwidth]{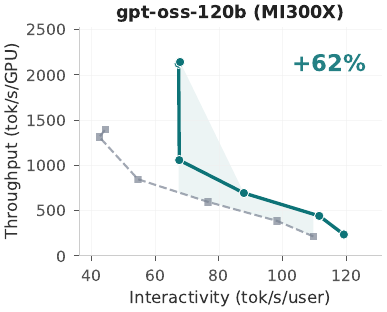}
  \hfill
  \includegraphics[width=0.32\textwidth]{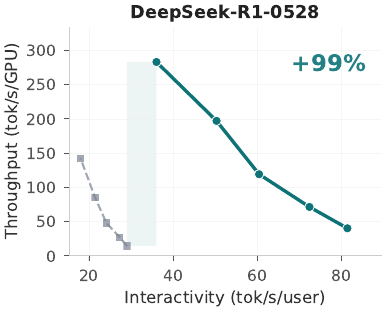}
  \hfill
  \includegraphics[width=0.32\textwidth]{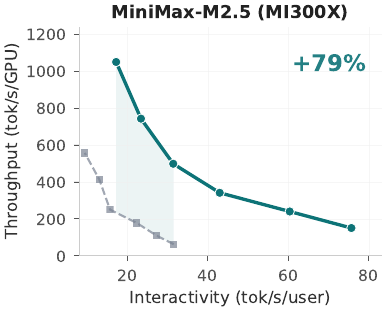}\\[4pt]
  \includegraphics[width=0.32\textwidth]{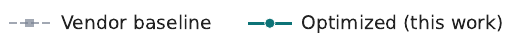}
  \caption{Throughput vs.\ interactivity on AMD MI300X for three production models. Same format as Figure~\ref{fig:pareto-mi355x}. No hardware-specific modifications to the agent or search policies.}
  \label{fig:pareto-mi300x}
\end{figure*}

Figure~\ref{fig:pareto-mi300x} shows the same evaluation on MI300X for three of the six models, using the same agent architecture and search policies with no hardware-specific modification. Arbor achieves +62\% on gpt-oss-120b, +79\% on MiniMax-M2.5, and +99\% on DeepSeek-R1-0528. The bottleneck distribution shifts substantially across hardware generations, yet the profiling-driven search adapts without manual intervention.

\section{Evaluation Metrics: Full Tables}
\label{app:eval-full}

Table~\ref{tab:eval-full} reports the evaluation framework metrics for all four policies on DSv4.1-Flash inference, corresponding to the radar chart in Figure~\ref{fig:radar}.

\begin{table}[h]
\caption{Evaluation framework metrics for DSv4.1-Flash inference (mean across runs per policy). Agent-driven baseline at 12\,h cutoff. All values are means across runs.
}
\label{tab:eval-full}
\vskip 0.1in
\begin{center}
\begin{small}
\begin{tabular}{@{}lcccc@{}}
\toprule
Metric & Bandit & Swarm & Sweep & AD@12h \\
\midrule
\multicolumn{5}{@{}l}{\textit{Efficiency}} \\
Gain / token cost & 0.50 & 0.19 & 0.44 & 0.23 \\
Gain / GPU-hr & 1.09\% & 0.99\% & 1.00\% & 0.30\% \\
Gain / iteration & 3.07\% & 4.08\% & 2.82\% & 0.69\% \\
Decisiveness & 0.94 & 0.99 & 0.97 & 0.77 \\
\midrule
\multicolumn{5}{@{}l}{\textit{Grounding}} \\
Integrity & 1.0 & 1.0 & 1.0 & 1.0 \\
Research rate & 1.00 & 1.00 & 1.00 & 1.00 \\
\midrule
\multicolumn{5}{@{}l}{\textit{Robustness}} \\
Recovery rate & 0.88 & 0.92 & 0.89 & 0.91 \\
Evidence on abandon & 1.00 & 1.00 & 0.91 & 1.00 \\
\midrule
\multicolumn{5}{@{}l}{\textit{Depth}} \\
Keep rate & 0.74 & 0.69 & 0.29 & 0.64 \\
Coverage & 2.5/5 & 5/5 & 1/5 & 4/5 \\
\bottomrule
\end{tabular}
\end{small}
\end{center}
\end{table}

\end{document}